\documentclass{article}
\usepackage{ijcai23}

\usepackage{times}
\usepackage{soul}
\usepackage{url}
\usepackage[hidelinks]{hyperref}
\usepackage[utf8]{inputenc}
\usepackage[small]{caption}
\usepackage{graphicx}
\usepackage{amsmath}
\usepackage{amsthm}
\usepackage{booktabs}
\usepackage{algorithm}
\usepackage{algorithmic}
\usepackage{multicol}
\usepackage{multirow}
\usepackage{amssymb}
\usepackage{makecell}
\usepackage{tablefootnote}
\usepackage{color}
\usepackage{pifont}
\usepackage{wrapfig}

\usepackage{amsmath,amsfonts,bm}

\def\eqref#1{equation~\ref{#1}}

\def\1{\bm{1}}

\def\vx{{\bm{x}}}

\def\mA{{\bm{A}}}

\def\mP{{\bm{P}}}

\def\mR{{\bm{R}}}
\def\mS{{\bm{S}}}

\def\mW{{\bm{W}}}
\def\mX{{\bm{X}}}

\DeclareMathAlphabet{\mathsfit}{\encodingdefault}{\sfdefault}{m}{sl}
\SetMathAlphabet{\mathsfit}{bold}{\encodingdefault}{\sfdefault}{bx}{n}

\def\gG{{\mathcal{G}}}

\def\gL{{\mathcal{L}}}

\newcommand{\R}{\mathbb{R}}

\DeclareMathOperator*{\argmin}{arg\,min}

\usepackage[switch]{lineno}
\DeclareMathOperator*{\fm}{FM}

\title{Prompt Federated Learning for Weather Forecasting: \\Toward Foundation Models on Meteorological Data}

\author{
Shengchao Chen
\and
Guodong Long\and
Tao Shen\And
Jing Jiang
\affiliations
Australian Artificial Intelligence Institute, FEIT, University of Technology Sydney\\
\emails
shengchao.chen.uts@gmail.com,
\{guodong.long, tao.shen, jing.jiang\}@uts.edu.au
}

\begin{document}

\maketitle

\begin{abstract}
To tackle the global climate challenge, it urgently needs to develop a collaborative platform for comprehensive weather forecasting on large-scale meteorological data. Despite urgency, heterogeneous meteorological sensors across countries and regions, inevitably causing multivariate heterogeneity and data exposure, become the main barrier. This paper develops a foundation model across regions capable of understanding complex meteorological data and providing weather forecasting. To relieve the data exposure concern across regions, a novel federated learning approach has been proposed to collaboratively learn a brand-new spatio-temporal Transformer-based foundation model across participants with heterogeneous meteorological data. Moreover, a novel prompt learning mechanism has been adopted to satisfy low-resourced sensors' communication and computational constraints. The effectiveness of the proposed method has been demonstrated on classical weather forecasting tasks using three meteorological datasets with multivariate time series.

\end{abstract}

\section{Introduction}
Climate change will significantly impact all regions; however, the specific effects will vary~\cite{kjellstrom2016heat}. Increasing global temperatures and melting ice will lead to alterations in sea levels, ocean currents, weather patterns, and cloud cover~\cite{hagemann2013climate}. To effectively tackle the challenge of global climate change, the implementation of a large-scale collaborative data-sharing platform is essential. Although this work is labor-intensive and demands a multitude of skilled experts, the utilization of machine learning techniques can enhance efficiency in addressing this problem. Nonetheless, the machine learning domain faces a  challenge when attempting to employ a centralized uniform model to serve all regions due to their heterogeneity. An effective solution involves pre-training a foundational model using extensive weather data and enabling each region to fine-tune the model using a relatively small data to enhance its ability to capture local weather patterns.
\begin{figure}[t]
    \centering
    \includegraphics[width=0.39\textwidth]{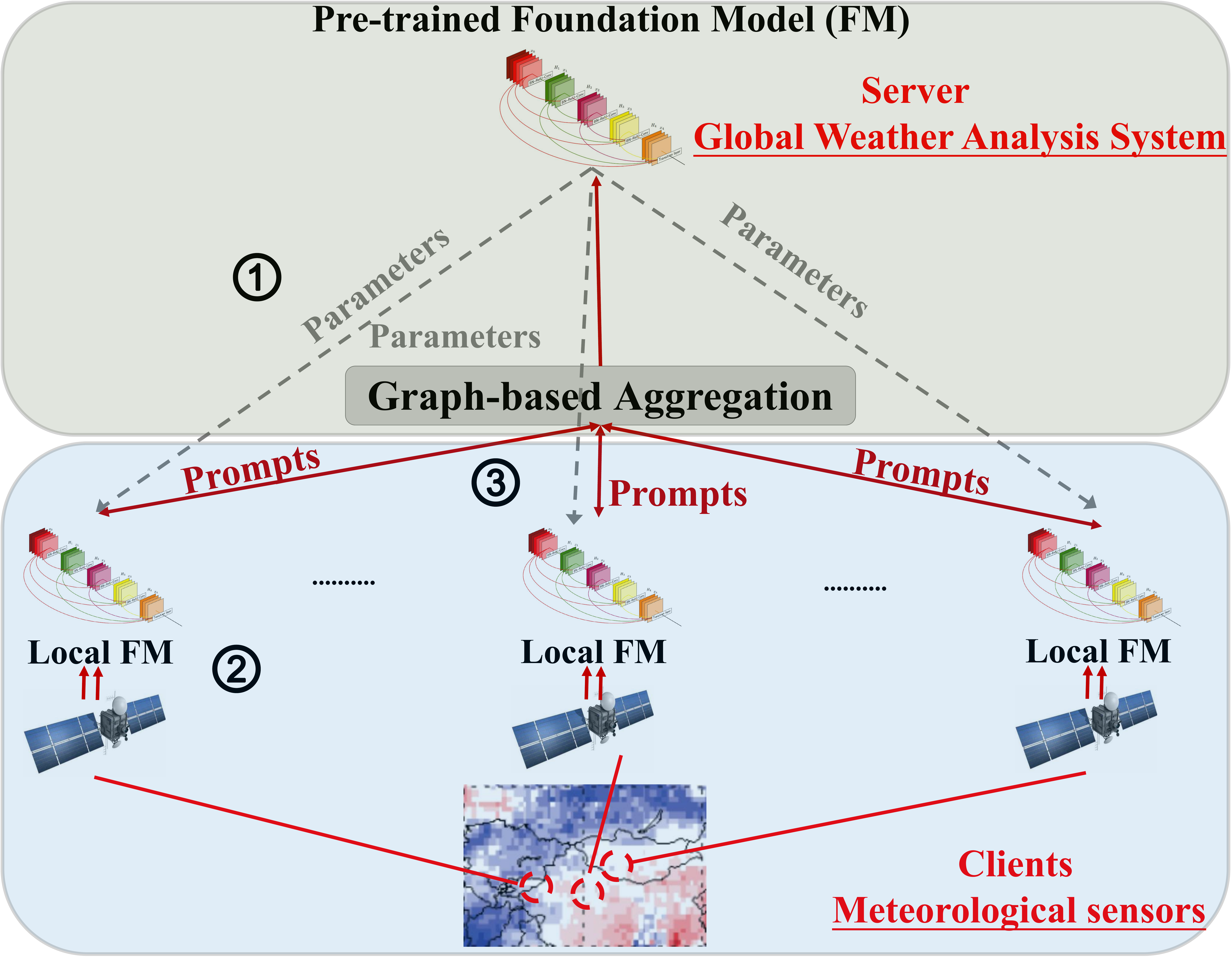}
    \caption{\small Our \textit{MetePFL} for weather forecasting. i) pre-trained FM initializes the local FM; ii) local FM trains using local data; iii) the server aggregates and transmits local prompts' parameters.}
    \label{Framework}
\end{figure}

Weather forecasting is a fundamental analytical task aimed at modeling the dynamic changes of weather on both a global and regional scale. Multi-sensor weather forecasting serves as a critical tool in mitigating the loss of human lives and property by providing early warnings for extreme weather events resulting from global climate change~\cite{chattopadhyay2020analog}. The objective of this approach is to capture potential correlations between multiple meteorological factors and the tendency of weather variations in order to gain a comprehensive understanding of specific regions. Unlike conventional time-series data, weather time-series data in meteorology are gathered from sensing devices distributed across diverse geographical locations~\cite{campbell2005weather}. 

One-step forecasting~\cite{chen2022dynamic,chen2023tempee}, empowered by recent advancements in deep learning, have garnered considerable attention due to their high efficiency. However, the performance of these strategies is hindered by the non-stationary nature of weather changes. This limitation arises from their reliance on fixed patterns derived from prior knowledge. To address this issue and capture temporal information, other studies~\cite{karevan2020transductive,alleon2020plumenet} suggest formulating the tasks as auto-regression problems. These studies utilize preceding time step variables to predict variables at the subsequent time step~\cite{chen2011comparison}. This technique is commonly implemented using RNN or Transformer models. The choice of these models is based on their superior performance in time series analysis~\cite{shi2015convolutional}.

However, previous works focus on using an uniform model to serve all regions regardless of heterogeneity. In contrast, foundation model (FM) represents a novel service architecture that aims to pre-train a large model with extensive data. Subsequently, this model can be fine-tuned for specific tasks using relevant data, such as weather forecasting in a particular region. The FM has the capability to capture common knowledge shared among multiple tasks or participants. Further refinements can then enhance its alignment with the specific requirements of a given task. The FM has demonstrated remarkable success in Natural Language Processing (NLP), exemplified by ChatGPT~\cite{schulman2022chatgpt}. Notably, recent progress in foundation models has been observed across diverse domains, including ViT~\cite{xu2022vitpose+}, BERT~\cite{yates2021pretrained}, and CLIP~\cite{radford2021learning}.

In contrast to existing FMs, training a FM on weather forecasting tasks must tackle the following challenges. First, sharing raw data across countries/regions will not be easy. Second, transmitting and processing the continuously collected data is a challenge for low-resourced sensors or devices. Third, real-time forecasting is critically important. 
In summary, we need a solution to tackle data security, communication, and computation efficiency issues and provide on-device decisions independently.

This paper will design a novel machine-learning approach to train foundation models on weather forecasting tasks. The model will be capable of understanding and constructing the complex spatiotemporal relationship of meteorological data to provide reliable analysis support on weather forecasting and global climate challenges.
Specifically, we propose a novel \textbf{Mete}orological \textbf{P}rompt \textbf{F}ederated \textbf{L}earning (\textit{MetePFL}) approach to collaboratively learn a Transformer-based foundation model (FM) across devices with multivariate time-series data (see Figure~\ref{Framework}). The \textit{MetePFL} only considers the model parameters exchange among devices rather than direct data sharing. Considering the low-resourced sensors' communication efficiency constraint, a brand-new prompt learning mechanism is introduced upon a pre-trained FM to comprehensively explore the correlation among weather-related variables while computing a few parameters.

Three weather forecasting datasets based on multivariate time series with multiple meteorological factors, i.e., precipitation, temperature, upstream, are leveraged to verify the effectiveness of \textit{MetePFL}.
The main contributions of this work are summarized in four-fold: 
\begin{itemize}
    \item This is the first work to explore a foundation model-based solution to enhance weather forecasting tasks towards a global scale. 
    
    \item The proposed prompt federated learning approach is a novel mechanism to collaboratively learns a foundation model for the applications with many satellite sites or stations across regions.
    
    \item A spatio-temporal prompt learning mechanism has been designed to efficiently tackle multivariable time series.

    \item Experiments on three real datasets have demonstrated the effectiveness and superior of our proposed solution. It is worth noting that we obtain excellent performance with only 2.38$\%$ of the model's parameters trained.
\end{itemize}

\section{Related Work}
\paragraph{Weather Forecasting.}
Weather forecasting plays a crucial role in the global climate analysis system. Conventional forecasting methods utilize numerical weather prediction (NWP) models, which incorporate physical constraints to simulate weather phenomena~\cite{bauer2015quiet}. However, with the emergence of data-driven approaches, weather forecasting has shifted towards approaches driven by data, such as ARIMA~\cite{chen2011comparison}, SVM~\cite{sapankevych2009time}, and NNs~\cite{voyant2012numerical}. While these basic models exhibit potential, they encounter difficulties in comprehending nonlinear temporal dynamics. Deep Learning methods, especially models based on Recurrent Neural Networks (RNNs), have exhibited promising outcomes in weather forecasting~\cite{shi2015convolutional}. Recently, Transformers have demonstrated superior performance compared to RNN-based models in time series analysis~\cite{bojesomo2021spatiotemporal}, thereby gaining popularity for weather-related tasks~\cite{chen2023tempee}. However, these models neglect the data exposure concerns when utilizing multi-sensor data for real-world tasks and modeling spatio-temporal correlations.

\paragraph{Foundation Model.} A fully supervised learning paradigm needs a large scale of data. The foundation model provides a practical solution for scenario-specific tasks, aiming to pre-train a model using extensive prior knowledge. The FM has widespread applications in NLP~\cite{yates2021pretrained,schulman2022chatgpt} and CV~\cite{xu2022vitpose+,radford2021learning}, providing an effective cross-task learning strategy. For example, ChatGPT~\cite{schulman2022chatgpt} can be used as a baseline model for researchers to fine-tune it to achieve more accurate responses for downstream tasks.

\paragraph{Federated Learning.}
Federated learning (FL) is a new learning paradigm to embody the collaborative training of models without requiring data exposure from each participant (e.g., meteorological sensors)~\cite{mcmahan2017communication,long2021federated,long2020federated,jiang2020decentralized}. Vanilla FL suffer from the heterogeneity of data and devices. Personalized FL aims to solve the problem by multiple techniques~\cite{tan2022federated1,chen2022personalized,wang2022fednoil,ma2022convergence,tan2021fedproto,zhang2023dual,li2023federated,long2023multi,li2019convergence,li2020federated,gao2022feddc} via training better personalized model for each client. However, these methods are not suitable for weather forecasting due to all parameters must be considered during communication so that hinder the real-time forecasting. pre-train-based strategy can mitigate the problem~\cite{tan2022federated} but can not explore the spatio-temporal correlations. Different from the above methods, this paper focuses on establishing a high-efficiency FL approach that provides analytical support to across-regional weather forecasting systems with heterogeneous meteorological data.

\paragraph{Prompt Learning.}
Prompt learning as a lightweight mechanism is widely used in NLP~\cite{li2021prefix,liu2021gpt}, which requires fewer parameters and is more adaptive than fine-tuned pre-trained models by represented by several prompt tuning strategies in different applications~\cite{zhou2022learning}.
Different from language data, understanding multidimensional correlation among multivariate data in the weather forecasting task is critical. However, the key point is often ignored by the federated prompt learning method~\cite{guo2022promptfl}.
The paper introduces a novel prompt mechanism within the FL framework based on pre-trained FM to explore the temporal dynamics and the potential correlation among clients while computing only a few parameters.

\section{Problem Formulation}
Given $N$ clients that possess individual local private datasets $D$, each client has a multivariate time series denoted as $ \mX_i \in \R^{m \times n}$. In this notation, each sample at a specific time step $t$ is represented as $ \vx_t\in \R^{1 \times n}$. Weather forecasting using multivariate time series can be defined as the process of utilizing historical values of all variables for a duration of $P$ periods to predict the values of a specific variable in the future over $Q$ periods, can be defined below:
\begin{small}
\begin{equation}
\left[ \vx_{t-P},  \vx_{t-P+1}, \cdots,  \vx_{t}\right] \stackrel{ f}{\longrightarrow}\left[ \vx_{t+1}^{\prime},  \vx_{t+2}^{\prime}, \cdots,  \vx_{t+Q}^{\prime}\right],
\end{equation}
\end{small}
where $ f$ is a learning system, and $ \vx_{t}^{\prime} \in  \R^{1 \times 1}$ is the value of the variable to be forecasting at the $t$-th time step.
Valida FL system aims to minimize the average loss of the global model $w$ on all clients' local dataset:
\begin{equation}
    F(w) \text{:}= \mathop{\arg\min}\limits_{w_1, w_2, ..., w_N}  \sum_{k=1}^{N} \frac{n_k}{n} F_k(w_k),
\end{equation}
where $n_k$ is the number of samples hold by the $k$-th client. $n$ is the number of samples held by all clients. $F_k(w_k)$ denotes the local objective function of $k$-th client. 
The distinguishing factor is that each client possesses a unique pattern, and sensors are deployed in specific locations, resulting in a statistically heterogeneous environment. To address this challenge, PFL is typically modeled as a bi-level optimization problem.
\begin{align}
    \notag F(v; w) \text{:} = \mathop{\arg\min}\limits_{\lbrace w_1, w_2, ..., w_N \rbrace, \lbrace v_1, v_2, ..., v_N \rbrace} \sum_{k=1}^{N} G_k(v_k, w), \\
    i.e. \quad G_k(v_k, w)  = \frac{n_k}{n} F_k(v_k) + \lambda R(v_k, w),
\end{align}
where each client hold a personalized model parameterized by $v_i$, $w$ denotes the global model. $R(\cdot)$ is the regularization term to control model update, via avoiding the local model updating be far away to the optimal global model. 

\section{Meteorological Prompt Federated Learning}
The framework of \textit{MetePFL} is depicted in Figure \ref{Framework}. In contrast to conventional Federated Learning (FL) where random global parameters are broadcasted to each client, \textit{MetePFL} employs a fixed FM, thereby reducing computation costs and improving performance without requiring extensive backpropagation. During each round, only the prompt parameters of the clients are taken into consideration. The \textit{MetePFL} framework consists of the Spatial-Temporal Prompt (STP) and the optimization process.

\subsection{Spatial-Temporal Prompt}
\begin{figure}[t]
    \centering\includegraphics[width=0.46\textwidth]{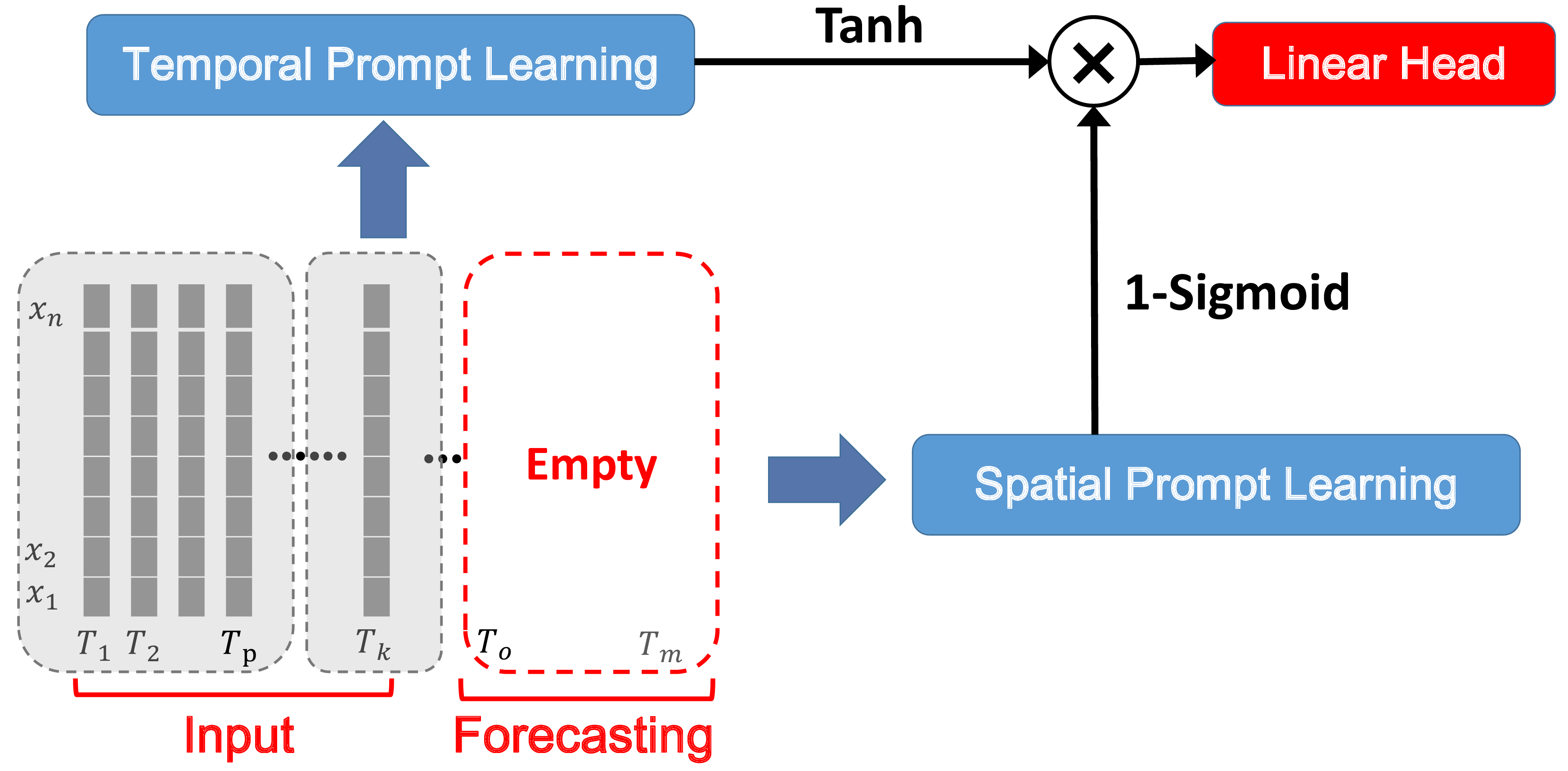}
    \caption{Schematic of Spatial-Temporal Prompt Learning.}
    \label{F2}
\end{figure}

The Spatial-Temporal Prompt (STP) shown in Figure \ref{F2} can be divided into Temporal prompt learning (TPL) and Spatial prompt learning (SPL). 

\begin{figure*}[t]
    \centering
    \includegraphics[width=0.85\textwidth]{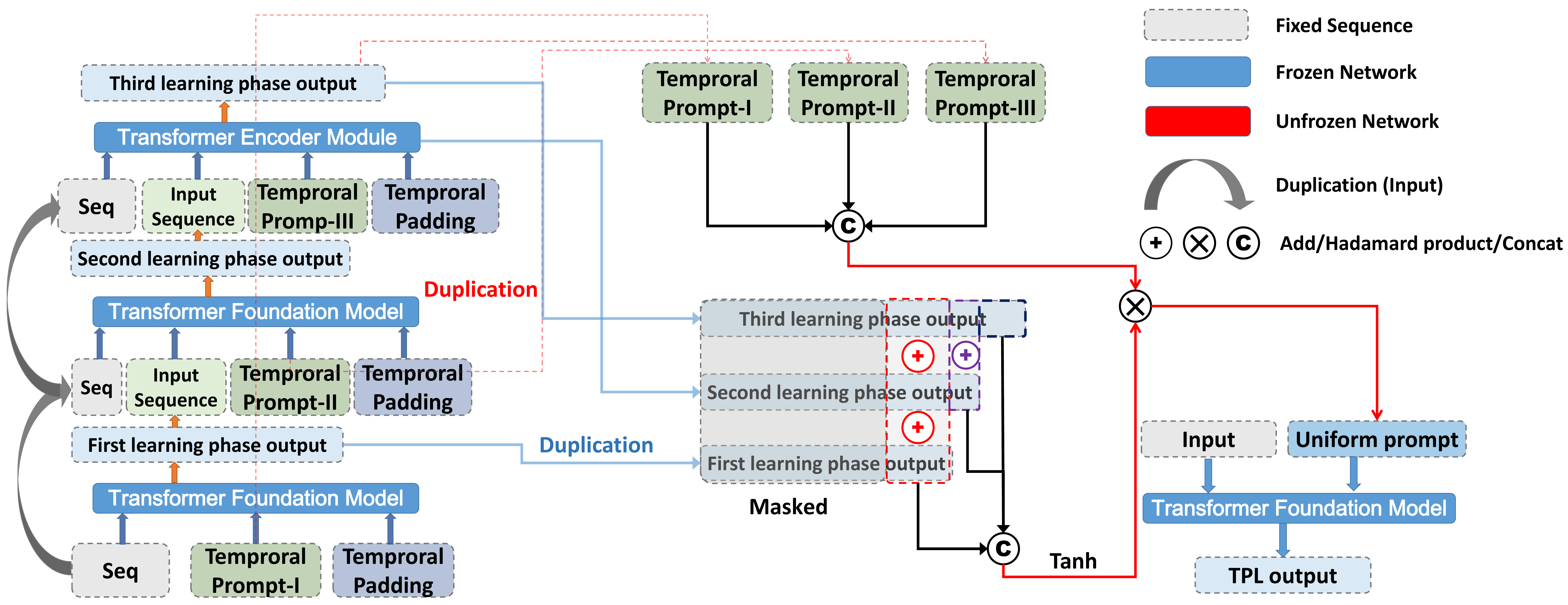}
    \caption{The learning strategy of Temporal Prompt Learning, which consists of four different learning phases.}
    \label{TPL}
\end{figure*}

\paragraph{Temporal prompt learning (TPL).}
To effectively understand the temporal dynamics under multivariate interactions, we use a multi-step incremental learning mechanism for learning the prompt parameters along the time dimension. The TPL comprise four phases, is shown in Figure~\ref{TPL}.

Given a time-series $ \mX \in  \R^{m \times n}$, the $m$ and $n$ represent the number of time steps and variables, respectively. Define a initial step $l$, the four temporal prompt format: (1) initial prompt $\hat{ \mP}$; (2) Temporal prompt-I: $ \mP_1$; (3) Temporal prompt-II: $ \mP_2 \in   \R^{2l \times n}$; (4) Temporal prompt-III: $ \mP_3 \in   \R^{2l \times n}$.

The first $p$ steps are fixed as $Seq \in  \R^{p \times n}$. In addition, the data within the $ p$ to $ k$ steps are spliced with the $\hat{\mP} \in  \R^{(k-p+l) \times n}$ to generate $ \mP_1$ by
\begin{equation}
    \Vert \mX^{(p \sim k)}, \hat{ \mP_T}\Vert \rightarrow  \mP_{T, 1},  \mP_{T, 1} \in   \R^{(2k-2p+l) \times n},
\end{equation}
where $\Vert \Vert$ denote the \textit{concat} operation. 
To enhance the accuracy of forecasting performance, we adopt a two-stage objective in the initial learning phase: "$\mX^{(0 \sim p)} \stackrel{1}{\longrightarrow} \mX^{(0 \sim k)} \stackrel{2}{\longrightarrow} \mX^{(0 \sim k+l)}$". During this phase, we encourage the FM to enhance prompt parameters associated with prior knowledge in $\mX^{(p \sim k)}$ in order to establish a foundation for the subsequent stage. In the next stage, the FM learns the remaining $\mP_{T, 1}$ parameters in $\mX^{(k \sim k+l)}$. This approach enables the FM to develop a comprehensive understanding of the relationships among variables across different time periods, rather than establishing rigid back-and-forth associations. The formulation of the initial learning phase is:
\begin{equation}
     \mR_1 = \fm([Seq, embed( \mP_{T,1})]),
\end{equation}
where the $embed(x)$ represents the learnable position encoder, and the $\fm$ is the pre-trained foundation model. 

Splicing discrete prompts to form a complete series is challenging because their parameters cannot match the FM simultaneously. To establish the relationship between two prompts, a dual-correct strategy is adopted in the second learning phase. This strategy encourages the FM to correct $\mP_{T,1}$ based on the previous $\mP_{T,1}$ while learning $\mP_{T,2}$. The objectives of the second learning phase are "$ \mX^{(0 \sim k+l)} \longrightarrow \mX^{(0 \sim k+3l)}$." This process can be formulated as mapping the input data from $ \mX^{(0 \sim k+l)}$ to $ \mX^{(0 \sim k+3l)}$ during the second phase.
\begin{small}
\begin{align}
    &\Vert \mP_{T,1},  \mP_{T,2}\Vert \rightarrow \mP_{T,2},  \mP_{T,2} \in  \R^{3l \times n},\\
    \notag & \mR_2 = \fm(\Vert Seq_2,  \mR^{p/2 \sim p}_1, embed( \mR^{p \sim p+l}_1), embed( \mP_{T,2})\Vert).
\end{align}
\end{small}

The objective of the third learning phase can be expressed as ``$ \mX^{(0 \sim k+3l)} \longrightarrow  \mX^{(0 \sim k+5l)}$'', for further correcting these previously learned prompts to improving the continuity of several prompts and providing smooth transitions between them. The third learning phase can be formulated as: 
\begin{small}
\begin{align}
    &\Vert \mP_{T,2},  \mP_{T,3}\Vert \rightarrow  \mP_{T,3},  \mP_{T,3} \in  \R^{4l \times n}, \\
    \notag & \mR_3 = \fm(\Vert Seq_3, \hat{ \mP} *  \mR^{p \sim (p+l)}_2, embed( \mP_{T,2}), embed( \mP_{T,3})\Vert).
\end{align}
\end{small}

To prevent the prompt parameters from being overly biased toward expressing short-term over long-term dependence, the final stage of learning intends to uniformly adjust these parameters. The corresponding values from the three previous learning phases are concatenated along the time dimension and then multiplied by the uniform prompt $ \mP_{T,w}$. The final learning can be expressed as follows.
\begin{align}
    \notag&\Vert \mP_{T,1},  \mP_{T,2},  \mP_{T,3}\Vert \rightarrow  \mP_{T,w},\\
    &\Vert \mP_{T,1} +  \mP_{T,2} +  \mP_{T,3},  \mP_{T,2} +  \mP_{T,3},  \mP_{T,3}\Vert \rightarrow \hat{ \mP_{T,w}},\\
    \notag & \mR_T = \fm(\Vert Seq, \frac{1}{m-k}\sum_{i=k}^m [tanh(\hat{ \mP_{T,w}}) *  \mP_{T,w}]\Vert).
\end{align}

\begin{figure*}[t]
    \centering
    \includegraphics[width=0.8\textwidth]{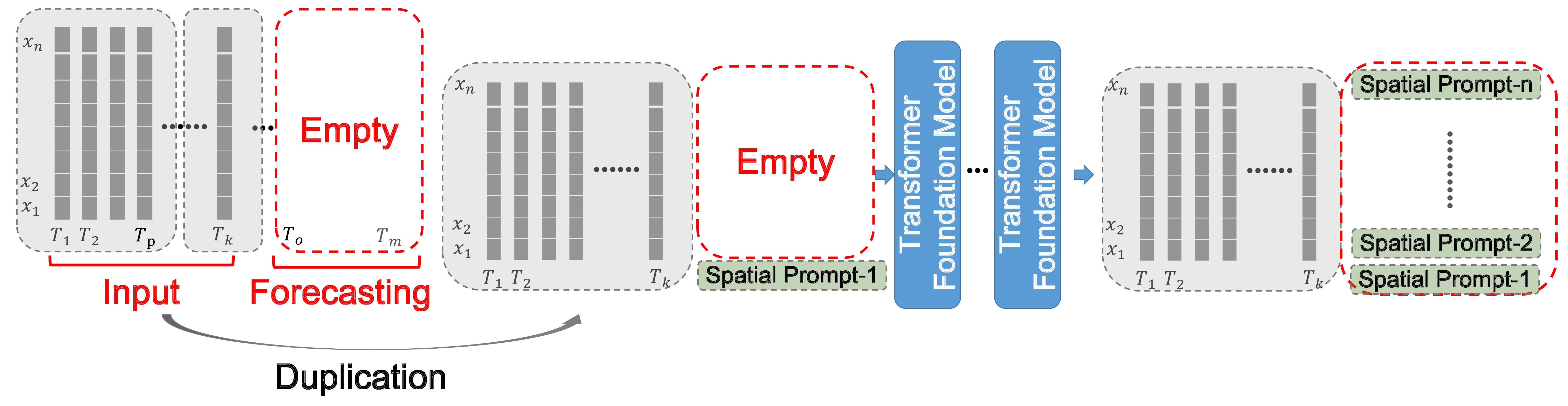}
    \caption{The learning strategy of Spatial Prompt Learning, the empty is a location in data where variables need forecasting.}
    \label{SPL}
\end{figure*}

\paragraph{Spatial Prompt Learning (SPL).}
We regard multiple meteorological factors within a specific space. This enables the establishment of correlations between these factors from a spatial perspective on the local client. The SPL are shown in Figure \ref{SPL}.

The trainable parameters that serve as spatial prompts can be represented as $\mathbf{P_S} \in \mathbb{R}^{(m-k) \times 1}$, where $m-k$ represents the length of the forecasting period. Prior to the initial learning phase, the first $p$ hours of the data are considered fixed and denoted as $Seq \in \mathbb{R}^{k \times n}$. Subsequently, $Seq$ are combined with the initial spatial prompts along the temporal dimension, and any gaps in the spatial dimension are filled with zero-valued parameters, resulting in $\hat{\mathbf{P_S}} \in \mathbb{R}^{(m-k) \times (n-1)}$. The first learning phase comprises:
\begin{equation}
\begin{aligned}
    &\Vert \mP_S, \hat{ \mP_S}\Vert \rightarrow  \mP_{S,1},\\
    & \mP'_{S,1} =  \mP_{S,1} * \fm(\Vert Seq,  \mS_1 \Vert).
\end{aligned}
\end{equation}

In learning spatial prompts for the $i^{th}$ variable, the $i^{th}$ learning phase of SPL can be formulated as follow:
\begin{equation}
\begin{aligned} 
    &\Vert \mP'_{S,i-1}, \hat{ \mP_S} \in  \R^{(m-k)\times(n-i)}\Vert \rightarrow \mP_{S,i},\\
    & \mP'_{S,i} =  \mP_{S,i} * \fm(\Vert Seq,  \mP_{S,i} \Vert).
\end{aligned}
\end{equation}
One advantage of SPL over auto-regression is its continuous correction of learned prompt parameters from the previous iteration. This correction utilizes all the previously predicted values, leading to improved forecast accuracy. Moreover, SPL improves the model's perception of correlations among multiple variables by considering adjacent variables together. Through multiple iterations of regression, SPL enhances the model's ability to account for these spatial correlations.

To prevent isolating the TPL and SPL effects within the STP, we combine the result of TPL and SPL empirically using a Gate operation. The output of the STP is formulated as $\mR = [(1 - sigmoid( \mR_S)) * tanh( \mR_T)] *  \mW$.

\subsection{Optimization of \textit{MetePFL}}
The optimization objective of \textit{MetePFL} is can formulated below according to the Eq. (3).
\begin{small}
\begin{align}
    \argmin_{\mP} \sum_{k=1}^{N} \frac{n_k}{n} [F_k(\lbrace \mP_k \rbrace) + \lambda R(\lbrace \mP_k \rbrace, \lbrace \mP^g\rbrace)] + \tau \gG(\mA)
\end{align}
\end{small}where $\lbrace\mP_k \rbrace$ is prompts including $\mP_S, \mP_T$, $\lbrace \mP^g \rbrace$ is the global prompt parameters, $\gG(\mA)$ is a regularization term used to represent the correlation among client according the adjacent matrix $\mA$. specifically, the term $F_k(\lbrace \mP \rbrace) + \lambda R(\lbrace \mP \rbrace, \lbrace \mP^g \rbrace)$ can be formulated as $\gL_k = \gL_{\text{MSE}} + \lambda \Vert \lbrace \mP_k \rbrace-  \lbrace \mP^g \rbrace \Vert_2^2$. The optimization objective on $i$-th client as:
\begin{equation}
    \begin{aligned}
    &\argmin_{\mA} \sum_{k=1}^{N} \lambda R(\lbrace \mP_k \rbrace, \lbrace \mP^g \rbrace) + \tau \gG(\mA) ,\\
    &s.t. \quad \mP^s  \in \argmin_{\lbrace \mP_k \rbrace} \mA_{j,i} S(\lbrace\mP_j\rbrace, \lbrace\mP_i \rbrace),\\
    &i.e. \quad \mP^g = G(\lbrace \mP^s_1 \rbrace, \lbrace \mP^s_2 \rbrace, ..., \lbrace \mP^s_N \rbrace),
    \end{aligned}
\end{equation}
where the $\mA \in \lbrace 0, 1\rbrace$, $S(\lbrace \mP_j \rbrace, \lbrace \mP_i \rbrace)$ is the similarity of prompt parameters of client $i$ and client $j$ measured by cosine or distance, $G(\cdot)$ is the average operation. During the optimization of \textit{MetePFL}, each client update their model via solving the local objective function as $\gL_k$ after them receive the fixed foundation model at first. Then each client upload their prompts $\mP$ rather than complete model to the server that conduct graph-based aggregation, which significantly reduce the communication overhead while exploring the potential correlations among clients. The aggregation including two steps: Graph Attention Network (GAT)~\cite{velivckovic2017graph}-based graph structure learning that explore the dynamic correlations among clients and Graph Convolution Network (GCN)~\cite{kipf2016semi} that utilized to parameters reconstruction using learned adjacent matrix $\mA$ and the prompt parameters uploaded by clients. The GCN automatically updates the parameters of each node by aggregating the models of its neighbors in the graph.

\begin{algorithm}[tbh]\small
\caption{\textit{MetePFL} algorithm.}
\begin{algorithmic}
\STATE $\mathbf{Initialized}$  $\mP_T$, $\mP_S$.
\STATE  $\mathbf{for}$ each communication round $t=0, 1, 2, \cdots, T$ $\mathbf{do}$ 
\STATE \quad\textbf{\textit{Local model initialize:}}
\STATE \quad $\mathbf{for}$ each client $i=0, 1, 2, ..., N$ in parallel $\mathbf{do}$
\STATE \quad\quad $0\leftarrow  \mP_T, \mP_S$
\STATE \quad\textbf{\textit{Local model update:}}
\STATE \quad $\mathbf{for}$ each client $i=0, 1, 2, \cdots, N$ in parallel $\mathbf{do}$
\STATE \quad\quad Update $\mP_T$, $\mP_S$ for $e$ local steps:
\STATE \quad\quad Train $\mP_T$, $\mP_S$ with loss function $\mathcal{L}$ in Eq.(11)
\STATE \quad $\mathbf{end}$ $\mathbf{for}$
\STATE Each selected client sends $\mP_T$, $\mP_S$ to the server
\STATE \textbf{\textit{Aggregation:}}
\STATE $ \mA \leftarrow Graph Generator(\lbrace \mP_1 \rbrace, \lbrace \mP_2 \rbrace,\cdots, \lbrace \mP_N \rbrace)$
\STATE $ \mA' \leftarrow GAT( \mA)$
\STATE Update $\mP$ for $r$ steps $GCN( \mA', \lbrace \mP_i \rbrace^N_{i=1})$:
\STATE \quad $\lbrace \mP_i^s \rbrace^N_{i=1} = \mA' \lbrace \mP_i \rbrace^N_{i=1}$
\STATE $\lbrace \mP_i^s \rbrace^N_{i=1} = \alpha \lbrace \mP_i^s \rbrace^N_{i=1} + (1 - \alpha) \lbrace \mP_i \rbrace^N_{i=1}$
\STATE Get Global Prompts $\lbrace \mP^g \rbrace \leftarrow G(\lbrace \mP_1^s \rbrace , \lbrace \mP_2^s \rbrace, ..., \lbrace \mP_N^s \rbrace)$
\STATE $\mathbf{end}$ $\mathbf{for}$
\end{algorithmic}
\label{alg1}
\end{algorithm}

\section{Experiments}

\paragraph{Baselines.}
We compare our \textit{MetePFL} with STGCN~\cite{yu2017spatio}, LSTM~\cite{graves2012long}, ConvLSTM~\cite{shi2015convolutional}, Transformer~\cite{zerveas2021transformer}, Informer~\cite{zhou2021informer}, Autoformer~\cite{wu2021autoformer}, and Fedformer~\cite{zhou2022fedformer}. The LSTM-based models have four layers. The Transformer consists of an eight-layer Encoder, while the Informer, Autoformer, and FEDformer consist of two encoders and a decoder. The Transforme models are trained by data-centric and FL setting, respectively.

\paragraph{Datasets.}
We compiled three multivariate time series datasets from NASA\footnote{https://disc.gsfc.nasa.gov/}, Average Precipitation (AvePRE), Surface Temperature (SurTEMP), and Surface Upstream (SurUPS) collected by 88, 525, and 238 devices, respectively. All three datasets cover the hour-by-hour variability of 12 different weather-related meteorological variables.

\paragraph{Experimental Setups.}
Models' input and output dimensions are $\mathcal{C}$ = 12 and $\mathcal{C}$ = 1, respectively. The dataset is split into training, validation, and testing in a 6:2:2 ratio. For pre-training, we use $2/3$ of the training set (i.e., $50\%$ of the entire dataset) for training and $1/6$ as the validation set (i.e., $10\%$ of the complete dataset) based on the above partition, following the pre-training strategy from Zerveas et al.~\cite{zerveas2021transformer}. For fine-tuning and prompt learning, the last $1/6$ of the training set is used for training, while the validation and test sets remain unchanged. In the federated training process, we set $k$ to control the number of clients participating in training per round, and we use $k$ = 0.1 and $k$ = 0.2 in the experiments. The forecasting uses 12 time steps in history ($\mathcal{P}$=12, i.e., the past twelve hours) to predict 15 time steps in the future ($\mathcal{Q}$=15, i.e., fifty hours in the future) with a time window length of 27 h. We set $l$ to 3, considering the validity time and trigger threshold of weather events.
\begin{table*}[tbh]
  \centering
  \resizebox{0.85\textwidth}{!}{
    \begin{tabular}{cccccccc}
    \toprule
    \multirow{2}[4]{*}{Model} & \multirow{2}[4]{*}{Temproal Encoding} & \multicolumn{2}{c}{AvePRE} & \multicolumn{2}{c}{SurTEMP} & \multicolumn{2}{c}{SurUPS} \\
     &    & MAE & RMSE & MAE & RMSE & MAE & RMSE \\
     \midrule
    STGCN & None & 0.331 & 0.815 & 0.196 & 0.257 & 0.298 & 0.399 \\
    ConvLSTM & None &  0.305  &  0.730  &  0.198  &  0.266  &  0.311  & 0.416 \\
    LSTM & None &  0.326  &  0.781  &  0.212  &  0.274  &  0.335  & 0.431 \\
    \midrule
    \multirow{2}[2]{*}{Transformer} & Learnable & 0.301 & 0.714 & 0.236 & 0.313 & 0.369 & 0.475 \\
       & Fixed & 0.332 & 0.744 & 0.239 & 0.320  & 0.351 & 0.449 \\
    \multirow{2}[2]{*}{FEDformer} & Learnable & 0.239 & 0.547 & 0.165 & 0.214 & 0.205 & 0.271 \\
       & Fixed & 0.248 & 0.564 & 0.165 & 0.216 & 0.201 & 0.264 \\
    \multirow{2}[2]{*}{Autoformer} & Learnable & 0.271 & 0.589 & 0.167 & 0.235 & 0.201 & 0.265 \\
       & Fixed & 0.269 & 0.590 & 0.176 & 0.228 & 0.212 & 0.279 \\
    \multirow{2}[2]{*}{Informer} & Learnable & 0.213 & 0.543 & 0.191 & 0.245  & 0.251 & 0.330  \\
       & Fixed & 0.216 & 0.547 & 0.193 & 0.251 & 0.240 & 0.311  \\
    \midrule
    Fed-Transformer & Learnable & 0.454/0.402  & 0.927/0.892  & 0.780/0.684 & 0.910/0.793 & 0.621/0.522 & 0.769/0.640 \\
    Fed-FEDformer & Learnable & 0.397/0.372   & 0.791/0.726   &  0.684/0.530  &  0.822/0.680  & 0.612/0.512 & 0.754/0.647 \\
    Fed-Autoformer & Learnable & 0.425/\underline{0.349} & \underline{0.784}/\underline{0.724} & 0.742/0.627 & 0.924/0.765 & \underline{0.578}/\underline{0.503} & \underline{0.715}/\underline{0.602} \\
    Fed-Informer & Learnable & \underline{0.385}/0.361 & 0.865/0.768 & \underline{0.647}/\underline{0.513} & \underline{0.790}/\underline{0.656} & 0.605/0.543 & 0.737/0.724 \\
    \midrule
    PromptFL-Transformer & Learnable & 0.427/0.389 & 0.828/0.786 & 0.683/0.612 & 0.824/0.741 & 0.603/0.519 & 0.766/0.641 \\
    \textit{MetePFL}* & Learnable & 0.389/0.376 & 0.631/0.626 & 0.592/0.522 & 0.611/0.597 & 0.584/0.485 & 0.721/0.610 \\
    \textit{MetePFL} & Learnable & \textbf{0.378}/\textbf{0.342} & \textbf{0.628}/\textbf{0.605} & \textbf{0.556}/\textbf{0.542} & \textbf{0.601}/\textbf{0.569} & \textbf{0.521}/\textbf{0.460} &  \textbf{0.642}/\textbf{0.589}\\
    \bottomrule
    \end{tabular}}
  \caption{Performance comparison of \textit{MetePFL} with baselines, the first eight models are trained from scratch using full parameters. Fed-Transformer refers to training the Transformer model from scratch in a federated learning (FL) setting, the last two models employ FL-based prompt learning methods with a pre-trained Transformer as the FM, the symbol $*$ indicates a FedAvg-based implementation, \underline{underline} means the optimal in FL full parameters training, \textbf{Bold} means the optimal in FL prompt learning strategy.}\label{OverallCom}
\end{table*}%
All models were trained on an NVIDIA Tesla V100 GPU using an initial learning rate of $1e^{-3}$ and a batch size of 128, ADAM~\cite{kingma2014adam} as the optimizer. The algorithm used in the FL of the Transformer-based network is FedAvg. The communication round was set to 20, $\alpha=0.5$, and the early stopping strategy was applied. Mean absolute error (MAE) and root mean absolute error (RMSE) as evaluation metrics. The code is avaliable at \textit{\url{https://github.com/shengchaochen82/MetePFL}}.

\begin{table*}[tbh]
  \centering
  \resizebox{0.80\textwidth}{!}{
    \begin{tabular}{cccccccc}
    \toprule
    \multirow{2}[4]{*}{FM} & \multirow{2}[4]{*}{Algorithm} & \multicolumn{2}{c}{AvePRE} & \multicolumn{2}{c}{SurTEMP} & \multicolumn{2}{c}{SurUPS} \\
\cmidrule{3-8}       &    & MAE & RMSE & MAE & RMSE & MAE & RMSE \\
    \midrule
    \multirow{4}[2]{*}{Transformer*} & FedAtt & 0.507/0.467 & 0.836/0.823 & 0.978/0.947 & 1.279/1.186 & 0.705/0.686 & 0.828/0.820 \\
       & FedProx & 0.567/0.531 & 0.845/0.827 & 0.922/0.901 & 1.141/1.102 & 0.688/0.672  &  0.814/0.810\\
       & Scaffold & 0.567/0.536 & 0.833/0.811 & 0.930/0.899 & 1.232/1.200 & 0.697/0.676   & 0.817/0.808 \\
       & FedAvg & 0.611/0.591 & 0.823/0.810 & 0.998/0.896 & 1.118/1.115 & 0.706/0.699 & 0.832/0.821 \\
    \midrule
    \multirow{5}[2]{*}{Transformer} & \textit{MetePFL} (FedAtt) & 0.383/0.357 & 0.735/0.618 & 0.576/\textbf{0.520} & 0.603/0.575 & \textbf{0.511}/0.482 & 0.642/0.610 \\
       & \textit{MetePFL} (FedProx) & 0.399/0.385 & 0.691/0.633 & 0.564/0.542 & 0.686/0.667 & 0.556/0.512 & 0.702/0.651 \\
       & \textit{MetePFL} (Scaffold) & 0.385/0.353 & 0.755/0.627 & 0.602/0.531  &  0.727/0.600  & 0.560/0.512 & 0.719/0.645 \\
       & \textit{MetePFL} (FedAvg) & 0.389/0.376 & 0.631/0.626 & 0.592/0.522 & 0.611/0.597 & 0.584/0.485 & 0.721/0.610 \\
       & \textit{MetePFL} & \textbf{0.378}/\textbf{0.342} & \textbf{0.628}/\textbf{0.605} & \textbf{0.556}/0.542 & \textbf{0.601}/\textbf{0.569} &  0.521/\textbf{0.460} &  \textbf{0.642}/\textbf{0.589}\\
    \bottomrule
    \end{tabular}}
   \caption{Performance comparison between fine-tuned Transformer and \textit{MetePFL} with different FL algorithm, $*$ implying that the model applies fine-tuning strategy, the \textit{MetePFL} ($\cdot$) means that the implementation based on other FL algorithm, \textbf{Bold} means the optimal results.}\label{PerformanCom}%
\end{table*}%

\subsection{Overall Comparison}
Table \ref{OverallCom} presents a performance comparison between our \textit{MetePFL} and baselines. Results indicate the superiority of the Transformer-based model over the LSTM-based model and STGCN when trained centrally across all three datasets. However, when utilized as a FM within the FL framework, the performance of the Transformer-based model is diminished compared to the trained FM. This reduction can be attributed to the  heterogeneity of weather data collected from multiple sensors. Notably, while the Transformer may be less effective than other similar models in centralized training, it demonstrates a significant performance advantage over Fed-FEDformer, Fed-Autoformer, and Fed-Informer when employed as an FM within the \textit{MetePFL}. This observation suggests that STP enhances the Transformer's capability to comprehend spatiotemporal data. Furthermore, our reliable FM and aggregation algorithms (Transformer and FedAvg) surpass PromptFL~\cite{guo2022promptfl}. This outcome provides additional validation of the effectiveness and superiority of our proposed \textit{MetePFL}.

To evaluate the effectiveness and superiority of \textit{MetePFL} over fine-tuning, we implement different \textit{MetePFL} version based on four popular FL algoirthms: FedAtt~\cite{jiang2020decentralized}, FedProx~\cite{sahu2018convergence}, Scaffold~\cite{karimireddy2020scaffold}, and FedAvg~\cite{mcmahan2017communication}. We maintained the prompt setting and compared the results against fine-tuning. The results are presented in Table~\ref{PerformanCom}. Our experiments reveal the following findings: (1) \textit{MetePFL} outperforms the fine-tuning method, highlighting the sensitivity of spatiotemporal data correlation to the STP approach; (2) the graph-based aggregation used in \textit{MetePFL} surpasses other algorithms, indicating its effectiveness in mitigating the negative impact of Non-IID on performance.

\begin{table*}[tbh]
  \centering
  \resizebox{0.85\textwidth}{!}{
    \begin{tabular}{cccccccc}
    \toprule
    Model & Strategy & \multicolumn{2}{c}{AvePRE} & \multicolumn{2}{c}{SurTEMP} & \multicolumn{2}{c}{SurUPS} \\
    \midrule
    \multirow{6}[4]{*}{Fed-Informer} & Fine-tuning (FedAvg) & 0.397/0.391 & 0.776/0.759 & 0.734/0.713 & 0.874/0.864 & 0.669/0.631 & 0.824/0.781 \\
       & Fine-tuning (FedAtt) & 0.403/0.378 & 0.780/0.724 & 0.950/0.899 & 1.057/1.00 & 0.686/0.675 & 0.807/0.793 \\
\cmidrule{2-8}  & PromptFL & 0.407/0.361 & 0.742/0.727 & 0.722/0.700 & 0.867/0.832 & 0.671/0.658 & 0.786/0.753 \\
& \textit{MetePFL} (FedAvg) & 0.382/\textbf{0.357} & 0.631/0.618 & \textbf{0.701}/0.692 & 0.840/0.812 & 0.698/0.650 & 0.796/\textbf{0.725} \\
       & \textit{MetePFL} (FedAtt) & 0.392/0.387 & 0.776/0.758 & 0.754/0.696 & 0.905/0.839 &  0.669/\textbf{0.644}  & 0.772/0.738 \\
       & \textit{MetePFL} & \textbf{0.363}/0.358 & \textbf{0.630}/\textbf{0.601} & 0.713/\textbf{0.677}  &  \textbf{0.838}/\textbf{0.800} &  \textbf{0.652}/\textbf{0.644}  & \textbf{0.755}/0.739\\
    \midrule
    \multirow{6}[4]{*}{Fed-Autoformer} & Fine-tuning (FedAvg) & 0.378/0.373 & 0.761/0.758 & 0.693/0.682 & 0.848/0.834 & 0.610/0.523 & 0.767/0.721 \\
       & Fine-tuning (FedAtt) & 0.372/0.355 & 0.761/0.754 & 0.706/0.683 & 0.848/0.839 & 0.598/0.551 & 0.742/0.698 \\
\cmidrule{2-8}  & PromptFL & \textbf{0.355}/0.348 & 0.759/0.753 & 0.672/0.659 & 0.826/0.807 & 0.584/0.543 &  0.724/0.678\\
& \textit{MetePFL} (FedAvg) & 0.364/0.348 & 0.780/0.731 & 0.674/\textbf{0.630} & 0.820/0.781  & 0.564/0.520 & 0.736/0.656 \\
       & \textit{MetePFL} (FedAtt) & 0.372/0.341 & 0.762/0.754 & 0.689/0.641 & 0.824/0.766 & 0.549/0.525  & 0.717/0.650 \\
       & \textit{MetePFL} & \textbf{0.355}/\textbf{0.334} & \textbf{0.750}/\textbf{0.719} &  \textbf{0.666}/\textbf{0.630}  &  \textbf{0.814}/\textbf{0.750} &  \textbf{0.547}/\textbf{0.516}  & \textbf{0.712}/\textbf{0.649}\\
    \bottomrule
    \end{tabular}}
  \caption{Performance comparison of pre-trained Fed-Informer and Fed-Autoformer with different learning strategies, \textit{MetePFL} (FedAvg) and \textit{MetePFL} (FedAtt) implies the FedAvg- and FedAtt-based implementations, respectively, \textbf{Bold} means the optimal performance.}\label{Performance8}%
\end{table*}%

\subsection{Framework Applicability}
To determine the applicability of \textit{MetePFL}, we replaced its FM with pre-trained Informer and Autoformer using the same pre-training strategy. Additionally, we used fine-tuning and PromptFL as reference strategies based on FedAvg and FedAtt. The comparison results as shown in Table \ref{Performance8}. Our proposed \textit{MetePFL} remains valid for other Transformer-based FMs and significantly outperforms fine-tuning and PromptFL. By comparing the performance of \textit{MetePFL} under different FL algorithm, we demonstrate the effectiveness of the graph-based aggregation once again.

\subsection{Parameter Utilization}

Table~\ref{Parameter} compares parameter utilization for different federated prompt learning strategies. PromptFL and \textit{MetePFL} are significantly more advantageous than regular training (train from scratch) and fine-tuning, with parameter utilization of 2.22$\%$ and 2.38$\%$, respectively - nearly 28$\%$ lower than fine-tuning. While PromptFL is better than \textit{MetePFL} in parameter utilization (-0.16$\%$), it performs nearly 20$\%$ worse than \textit{MetePFL} (see Table~\ref{OverallCom}). Despite considering only 2.38$\%$ parameters, \textit{MetePFL} achieves excellent performance and significantly improves inter-device communication efficiency.

\begin{table}[H]
  \centering
   \resizebox{0.49\textwidth}{!}{
    \begin{tabular}{cccc}
    \toprule
    Strategy & $\#$ of Total Param & $\#$ of Training Param & $\#$ of Participation Param \\
    \midrule
    FL-Regular & 3,229,857 & 3,229,857 & 100$\%$ \\
    FL-Fine Tuning & 3,288,886 & 109,854 & 30.37$\%$ \\
    PromptFL & 3,250,867 & 71,835 & 2.22$\%$ \\
    \textit{MetePFL} (Ours) & 3,258,547 & 77,595 & 2.38$\%$ \\
    \bottomrule
    \end{tabular}}
  \caption{Comparison of parameter utilization of \textit{MetePFL}.}\label{Parameter}%
\end{table}%

\subsection{Ablation Study}
\begin{table}[tbh]
      \centering
  \resizebox{0.45\textwidth}{!}{
    \begin{tabular}{cccc}
    \toprule
    Model & Strategy & MAE & RMSE \\
    \midrule
    \multirow{4}[2]{*}{Encoder-only} & Train from scratch & 0.332 & 0.744 \\
       & Train from scratch \& STP & 0.300 & 0.689 \\
       & Pre-train & 0.295 & 0.668 \\
       & Pre-train \& STP & 0.267 & 0.571 \\
    \midrule
    Decoder-only & Train from scratch & 0.304 & 0.724 \\
    \bottomrule
    \end{tabular}}
  \caption{\small Comparison of Encoder/Decoder-only Transformer under different strategies in general scenarios based on AvePRE.}\label{Ablation1}%
\end{table}
To evaluate the effectiveness of STP, we conducted ablation studies under general scenarios rather than under the FL framework. The results are presented in Table~\ref{Ablation1}. The Encoder-only and Decoder-only Transformer underwent pre-training with one-step forecasting and auto-regressive mechanisms similar to previous experiments. The Encoder-only Transformer trained with STP outperformed the Decoder-only Transformer, indicating the superior learning mechanism of the proposed STP over conventional auto-regression. Moreover, the performance gap between the pre-trained Encoder with and without STP confirms the efficacy of the proposed STP in general scenarios.

\begin{table}[tbh]
  \centering
  \resizebox{0.32\textwidth}{!}{
    \begin{tabular}{cccccc}
    \toprule
    TPL & SPL & Gate & PE & MAE & RMSE \\
    \midrule
    \textit{w}  & \textit{w}  & \textit{w/o} & \textit{w/o} & 0.301 & 0.608 \\
    \textit{w}  & \textit{w}  & \textit{w}  & \textit{w/o} & 0.267 & 0.571 \\
    \textit{w/o} & \textit{w}  & \textit{w/o} & \textit{w/o} & 0.284 & 0.580 \\
    \textit{w}  & \textit{w/o} & \textit{w/o} & \textit{w/o} & 0.299 & 0.613 \\
    \textit{w/o} & \textit{w/o} & \textit{w/o} & \textit{w}  & 0.284 & 0.617 \\
    \bottomrule
    \end{tabular}}
  \caption{Ablation results of the proposed STP in Encoder-only Transformer based on the AvePRE in general scenarios.}\label{Ablation2}%
\end{table}%

The effectiveness of SPL, Gate, and their advantage over position-aware embedding (PE) was verified using the Encoder-only Transformer. The results are presented in Table~\ref{Ablation2}. SPL demonstrates a greater improvement compared to PE. Additionally, SPL enhances the model's performance, while the model without Gate experiences a significant decline. TPL does not perform as effectively as desired in generic scenarios. Consequently, we conducted ablation experiments in the FL setting, as shown in Table~\ref{T7}. The results indicate that both TPL and SPL enhance the model's forecasting performance. In conclusion, the effectiveness of the Gate operation is demonstrated, and both TPL and SPL can improve the model's performance in the FL setting.

\begin{table}[tbh]
  \centering
    \begin{tabular}{ccccc}
    \toprule
    TPL & SPL & Gate & MAE &  RMSE \\
    \midrule
    \textit{w}  & \textit{w} & \textit{w} & 0.378&0.628 \\
    \textit{w/o} & \textit{w} & \textit{w/o} & 0.407&0.722 \\
    \textit{w}  & \textit{w/o} & \textit{w/o} & 0.415&0.740 \\
    \bottomrule
    \end{tabular}
  \caption{Ablations on FL setting. }
  \label{T7}
\end{table}
\vspace{-12pt}
\section{Conclusion}
This paper proposes a novel machine learning approach to train foundation models for weather forecasting tasks, capable of capturing the spatiotemporal relationships of meteorological data based on multivariate time series. To enhance the performance while keeping data secure and reducing communication overhead, we introduce a prompt learning mechanism based on the fixed foundation model within the FL framework. Additionally, we utilize a graph-based approach to mitigate the impact of data heterogeneity on model effectiveness. Extensive experiments on three real-world weather datasets confirm the effectiveness of our proposed method.

\clearpage
\bibliographystyle{named}
\bibliography{ijcai23}

\end{document}